\pdfoutput=1

\documentclass[11pt]{article}

\usepackage{EMNLP2022}

\usepackage{times}
\usepackage{latexsym}

\usepackage{xcolor}
\usepackage{times}
\usepackage{float}
\usepackage{helvet}
\usepackage{amsmath}
\usepackage{amssymb}
\usepackage{tablefootnote}
\usepackage{bbding}
\usepackage{pifont}
\usepackage{flushend}

\usepackage{multirow}

\usepackage{wasysym}
\usepackage{subcaption}
\usepackage{comment}
\usepackage{graphicx}
\newsavebox{\largestimage}
\usepackage{wrapfig}

\usepackage{array}
\usepackage{makecell}

\usepackage{booktabs}

\usepackage[T1]{fontenc}

\usepackage[utf8]{inputenc}

\usepackage{microtype}

\usepackage{inconsolata}

\title{Linguistic Elements of Engaging Customer Service\\Discourse on Social Media}

\author{Sonam Singh$^1$ and Anthony Rios$^{2}$ \\
  $^1$Department of Marketing\\
   $^2$Department of Information Systems and Cyber Security\\
  University of Texas at San Antonio \\
  \texttt{\{sonam.singh, anthony.rios\}@utsa}}

\begin{document}
\maketitle
\begin{abstract}
Customers are rapidly turning to social media for customer support. While brand agents on these platforms are motivated and well-intentioned to help and engage with customers, their efforts are often ignored if their initial response to the customer does not match a specific tone, style, or topic the customer is aiming to receive. The length of a conversation can reflect the effort and quality of the initial response made by a brand toward collaborating and helping consumers, even when the overall sentiment of the conversation might not be very positive. Thus, through this study, we aim to bridge this critical gap in the existing literature by analyzing language's content and stylistic aspects such as expressed empathy, psycho-linguistic features, dialogue tags, and metrics for quantifying personalization of the utterances that can influence the engagement of an interaction. This paper demonstrates that we can predict engagement using initial customer and brand posts. 
\end{abstract} 

\section{Introduction}

Providing quality customer service on social media has become a priority for most companies (brands) today. A simple customer-brand interaction started by a moment of annoyance can be relieved when the brand resolves the issue in a public display of exceptional customer service. According to Forbes, companies that use Twitter as a social care channel have seen a 19 percent increase in customer satisfaction~\cite{Forbes}. Furthermore, customers are rapidly turning to social media for customer support; Research from JD Power finds that approximately 67 percent of consumers now tap networks like Twitter and Facebook for customer service~\cite{Power2013}. While providing timely and stellar service has its advantages, engaging in a collaborative dialogue with its customers also leads to mutual trust and transparency according to social customer relations management (SCRM) theories ~\cite{yahav2020journey}. SCRM  has been documented as a core business strategy~\cite{woodcock2011social}. SCRM refers to ``\textit{a philosophy and a business strategy, supported by a technology platform, business rules, processes, and social characteristics, designed to engage the customer in a collaborative conversation to provide mutually beneficial value in a trusted and transparent business environment}’’~\cite{greenberg2010crm}. While brand agents on social platforms are typically motivated and well-intentioned to help engage with customers, their efforts are often ignored if their initial response to the customer does not match a specific tone, style, or topic the customer is aiming to receive. Hence, we explore what textual elements of a brand’s response are predictive of customer engagement.

While there has been substantial research on customer service on social platforms, a majority has predominantly addressed issues such as timely response~\cite{xu2017new}, timely transfer from a bot to human ~\cite{liu2020time}, improving bot performance~\cite{adam2020ai,folstad2019conversational,xu2017new,hu2018touch}, improving dialogue act prediction~\cite{oraby2017may,bhuiyan2018don}, and managing customer sentiment of users on the platform ~\cite{mousavi2020voice}. A few studies have also examined the tone and emotional content of the customer service chats~\cite{hu2018touch,herzig-etal-2016-classifying,oraby2017may}. However, more subtle and integral stylistic aspects of language, such as expressed empathy, psycho-linguistic features (e.g., time orientation), and the level of personalization of the responses, have received little attention, particularly when analyzed for engagement metrics. For the few studies that have looked at subtle stylistic aspects of language in customer service settings~\cite{clark2013empathy,wieseke2012role}, the studies were more lab-based in synchronous, face-to-face, or call settings and may not translate to asynchronous, text-based contexts. Additionally, only emotional aspects of empathy were considered. Yet, similar to Face-to-Face communication, text-only communication also contains many subtle, and not so subtle, social cues.~\cite{jacobson1999impression,hancock2001impression,bargh2004internet,rouse2003exploring}. Broadly, there are two dimensions of language that can provide information about a conversation: language \textit{content} and \textit{style}. The \textit{content} of a conversation indicates the general topics being discussed, along with the relevant actors, objects, and actions mentioned in the text. Conversational \textit{style} reflects how it is said~\cite{pennebaker2011using}. Thus, a text-based response can be examined for its content and its style.

Language can be viewed as a fingerprint or signature~\cite{pennebaker2011using}. Besides reflecting information about the people, organizations, or the society that created it, the text also impacts the attitudes, behavior, and choices of the audience that consume it~\cite{berger2020uniting}. For example, the language of the response from a brand agent can assure and calm a consumer, infuriate them, or make a customer anxious. While language certainly reflects something about that writer (e.g., their personality, how they felt that day, and how they feel towards someone or something), the language also impacts the people who receive it~\cite{packard2021concrete,packard2018m}. It can influence customer attitudes toward the brand, influence future purchases, or affect what customers share about the interaction~\cite{berger2020uniting}. 

Overall, in this paper, we aim to examine how the initial query from a customer and the initial response from a brand’s agent impact the engagement of interaction on social media. However, rather than focusing just on the textual content of a conversation alone, we also examine the language style through the use of cognitive and emotional expressed empathy (e.g., emotional reactions, interpretations, explorations)~\cite{sharma2020computational}, psycho-linguistic (LIWC) language features (e.g., time orientation, tone)~\cite{pennebaker2015development},dialogue-tags, and novel use of perplexity as a metric of personalization of the responses~\cite{brown1992estimate,heafield-2011-kenlm}. Overall, to the best of our knowledge, we make a first attempt at examining a comprehensive set of content and style features from customer service conversations to understand their impact on \textit{engaging conversations} between brand agents and customers. In addition, this is the first study to analyze customer engagement as a measure of the effort of brand agents to provide customer support beyond positive sentiment. Moreover, we also build a prediction model to demonstrate the predictive capability of these stylistic features. We show it is possible to predict the likelihood of its engagement from the first response from a brand agent.

\section{Related Work}
Given the importance of customer service there exists a substantial body of research addressing different challenges in this area. A body of researchers focuses on improving chatbots. \citet{adam2020ai} build upon social response theory and anthropomorphic design cues. They find artificial agents can be the source of persuasive messages. However, the degree to which humans comply with artificial social agents depends on the techniques applied during human-chatbot communication. In contrast, \citet{xu2017new} designed a new customer service chatbot system that outperformed traditional information retrieval system approaches based on both human judgments and an automatic evaluation metric. \citet{hu2018touch} examine the role of tone and find that tone-aware chatbot generates as appropriate responses to customer requests as human agents. More importantly, the tone-aware chatbot is perceived to be even more empathetic than human agents. \citet{folstad2019conversational} emphasize the repair mechanisms (methods to fix bad chatbot responses) and find that chat-bots expressing uncertainty are bad in the customer service setting. Thus,  \citet{folstad2019conversational}  develop a method to suggest likely alternatives in cases where confidence falls below a certain threshold. 

Another stream of research examines the role of emotions and sentiment in service responses. For example, \citet{zhang2011effects} find that emotional text positively impacts customers' perceptions of service agents. Service agents who use emotional text during an online service encounter were perceived to be more social. \citet{guercini2014uncovering} identify elements of customer service-related discussions that provide positive experiences to customers in the context of the airline industry. They find that positive sentiments were linked mostly to online and mobile check-in services, favorable prices, and flight experiences. Negative sentiments revealed problems with the usability of companies’ websites, flight delays, and lost luggage. Evidence of delightful experiences was recorded among services provided in airport lounges. \citet{mousavi2020voice} explores the factors and external events that can influence the effectiveness of customer care. They focus on the telecom industry and find that the quality of digital customer care that customers expect varies across brands. For instance, customers of higher priced firms (e.g., Verizon and AT\&T) expect better customer care. Different Firms provide different levels of customer service and seemingly unrelated events (e.g., signing an exclusive contract with a celebrity) can impact digital customer care. 

There is also research that focuses on ``dialogue acts''---identifying utterances in a dialogue that perform a customer service-specific function.  For instance, \citet{herzig-etal-2016-classifying} study how agent responses should be tailored to the detected emotional response in customers, in order to improve the quality of service agents can provide. They demonstrate that dialogue features (e.g., dialogue acts/topics) can significantly improve the detection of emotions in social media customer service dialogues and help predict emotional techniques used by customer service agents. Similarly, \citet{bhuiyan2018don} develop a novel method for dialogue act identification in customer service conversations, finding that handling negation leads to better performance. \citet{oraby2017may} use 800 twitter conversations and develop a taxonomy of fine-grained ``dialogue acts'' frequently observed in customer service. Example dialogue acts include, but are not limited to, complaints, requests for information, and apologies. Moreover, \citet{oraby2017may} show that dialogue act patterns are predictive of customer service interaction outcomes. While their outcome analysis is similar to our study (i.e., predicting successful conversations), it differs in the ultimate analysis and goals. Specifically, they focus on function rather than language style. For example, their work can guide service representatives to ask a yes-no question, provide a statement, or ask for information. In contrast, in this paper, we focus on looking at specific language features. Our work can guide \textit{how} the customer representative should respond (e.g., provide a more specific response), which is different than describing \textit{what} they should do (e.g., ask for more information). 

Finally, closely related to our study, there are a range of studies examining the different aspects of language and its impact on customer service. For example, \citet{packard2021concrete} study linguistic correctness---the tangibility, specificity, or imaginability of words employees use when speaking to customers. They find customers are more satisfied, willing to purchase and purchase more when employees speak to them concretely. \citet{clark2013empathy} study the nature and value of empathetic communication in call center dyads. They find that affective expressions, such as ``\textit{I'm sorry},'' were less effectual, but attentive and cognitive responses could cause highly positive responses, although the customers' need for them varied substantially. \citet{wieseke2012role} find that customer empathy strengthens the positive effect of employee empathy on customer satisfaction, leading to more ``symbiotic interactions.'' The major difference between the prior studies and ours is that they study empathy in lab-based settings (i.e., real-world interactions) and not text conversations on social media. Furthermore, we correlate language to consumer engagement, which is missing in prior work.

\section{Data}

\begin{figure*}[t]
    \centering
    \includegraphics[width=\linewidth]{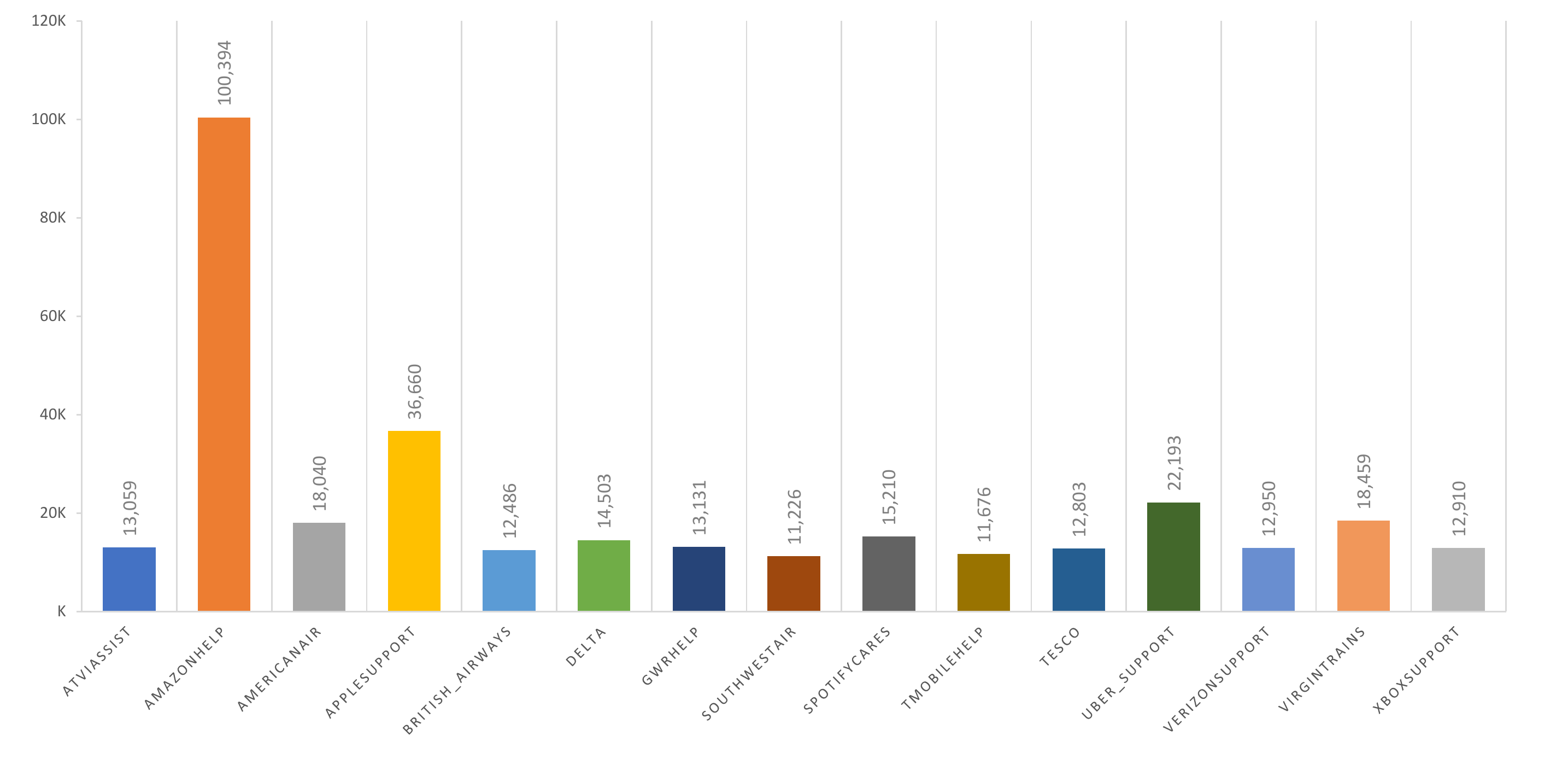}
    

    \caption{Plot of the number of tweets for Brands with at least 10K tweets in our dataset.\vspace{0em}}
    \label{fig:datadesc}
\end{figure*}

Our data set consists of customer service-related queries and brand responses from their Twitter service handles. We start with two million tweets (2,013,577) spanning over 789K conversations between 108 brands and 667,738 customers. Some brands have as few tweets as 107 conversations (e.g., OfficeSupport), while other brands such as Amazon and Apple have 100K and 36K conversations, respectively. Figure~\ref{fig:datadesc} plots the brands that have more than 10K tweets. The average number of tweets per conversation is 4.15 and the average number of words per tweet is 18.52. The length of these conversations varies substantially, while some are as short as one round (user tweet$\rightarrow$brand tweet$\rightarrow$end) others are as long as ten rounds. We measure \textit{engagement} by counting the number of brand$\rightarrow$customer interactions are made. For instance, if the customer writes a question, a brand responds, and then the customer never responds again, then that would have an engagement count of zero. If the customer responds once, then the engagement count is one.
\begin{table}[t]
\centering
\resizebox{0.8\linewidth}{!}{%
\begin{tabular}{@{}lll@{}}
\toprule
                      & \textbf{Train}  & \textbf{Test}  \\ \midrule
Total   Conversations             &    472,412    &    317,348   \\
Engaging              & 134,650 & 44,796 \\
Average Convo. Length & 4.15   & 4.14  \\
Max Convo. Length     & 604\footnotemark   & 432   \\ \bottomrule
\end{tabular}%
}
\caption{Dataset Statistics}
\label{tab:statss}
\end{table}

\footnotetext{Note that this includes multi-party conversations, the original user only replied four times in the longest conversation} 
To understand engagement better, we manually analyzed 500 conversations with more than 1 round (user$\rightarrow$brand$\rightarrow$user...) to understand the nature of these conversations. We found that 85\% of these conversations were putting substantial personalized effort towards trying to resolve customer issues, 1\% were appreciations from customers, and the remaining 14\% represented customers' frustrations/anger with a poor experience. Even among the 14\% of the conversations that expressed anger initially, we found that the brand agent showed effort towards helping the customer when the customer actively engaged with them. Thus, we find that the length of a conversation can reflect the effort and quality of the initial response made by a brand toward collaborating and helping consumers. While many of these conversations might get resolved offline and reflect neutral sentiment, or limited engagement, on social platforms, the conversation size can provide a helpful signal in finding conversation-related characteristics that signal quality brand responses. Hence, we are able to use this metric of engagement to better understand which language factors lead to it. 
\begin{figure*}
\centering
\savebox{\largestimage}{\includegraphics[width=.5\linewidth]{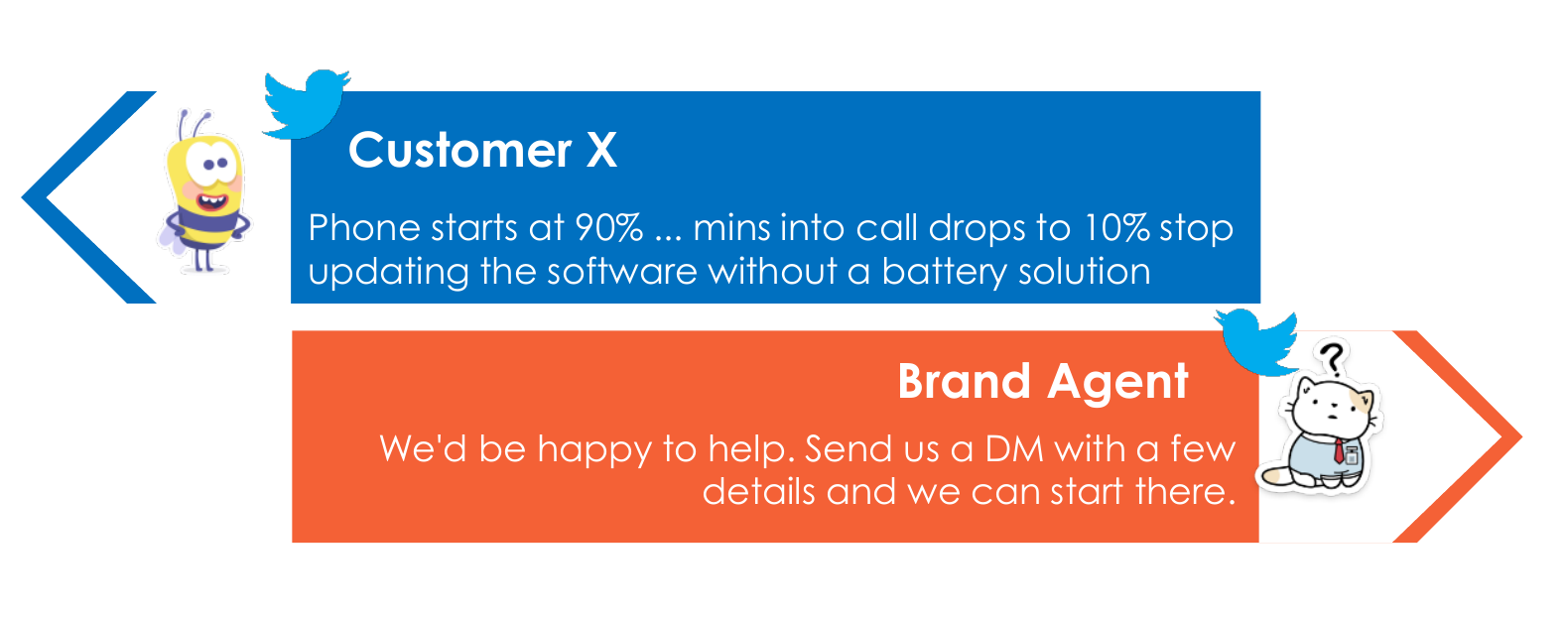}}%
\begin{subfigure}{.48\textwidth}
  \centering
    \usebox{\largestimage} \vspace{-.25em} 
    \caption{Non-Engaging Interaction}
    \label{fig:nonengaging}
\end{subfigure}%
\begin{subfigure}{.48\textwidth}
  \centering
  \raisebox{\dimexpr.5\ht\largestimage-.37\height}{%
    \includegraphics[width=.95\linewidth]{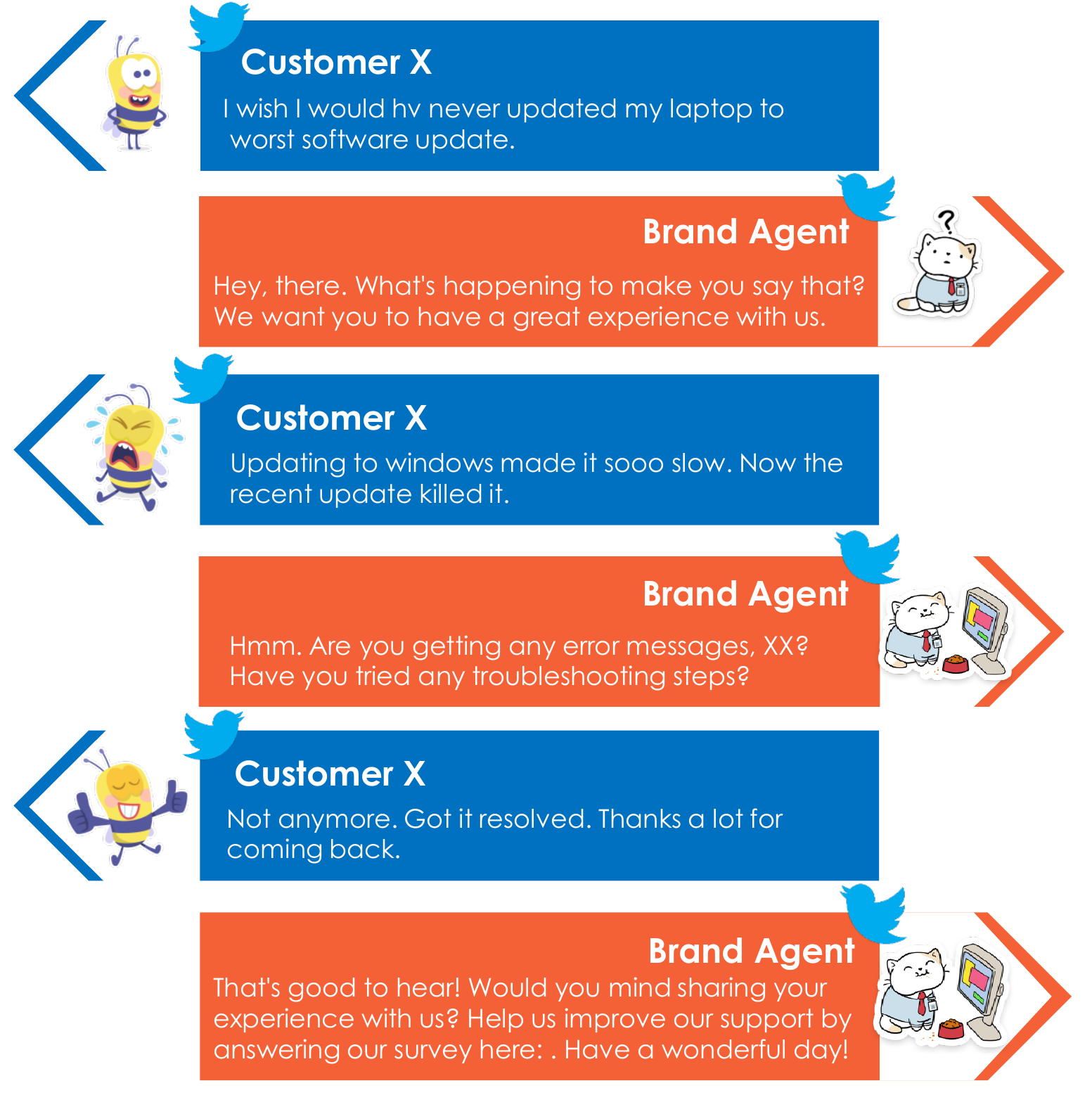}}\vspace{-1.em} 
    \caption{Engaging Interaction}
    \label{fig:engage}
\end{subfigure}%
\caption{Example of engaging and non-engaging interactions. The non-engaging interaction has a single customer response (length = 1) and the engaging conversation has multiple customer responses (length = 3).}
\end{figure*}
We operationalize the engagement in two forms. First, we consider the length of the conversation, for example, Figure~\ref{fig:engage} has a length of three. The length of the conversation represents the effort brand agents put in to resolve customer query. Second, we also construct a binary ``Engagement Indicator'' outcome variable which is ``engaging'' when the discourse has more than one round (user$\rightarrow$brand$\rightarrow$user$\rightarrow$..), and ``not-engaging'' when it ends in one round (user$\rightarrow$brand$\rightarrow$end). Figure~\ref{fig:engage} and Figure~\ref{fig:nonengaging} provide a sample example of a engaging vs. not engaging conversation.  Table~\ref{tab:statss} provides a summary of the dataset statistics.

\section{Method}
Our main research question is to understand how the content and stylistic features of the text influence the engagement of interaction in social media discourse. Thus, our methodology involves five major steps. In Step 1, we collect and understand data to identify engagement through the length. For Step 2, we generate the stylistic metrics to cover empathy, personalization/novelty of a response, dialogue tags, and general psycho-linguistic features. We also include content-based features. For Step 3, we operationalize engagement as an Engagement Indicator (``engaged'', if the length is greater than one or ``not engaged'' otherwise). Step 4 involves machine learning experiments. 
We start with all two million tweets (2,013,577) spanning over 789K conversations between 108 brands and 667,738 customers. We then randomly split (60:40) the entire data set into training and test sets (train size- 472,512; test size-  317,348).  For Step 5 we report the most predictive items for  all features. The details of the feature generation process are described in the following subsections.

\subsection{Content Features}

We explore two types of content features: bag-of-word features and neural content features generated using RoBERTa~\cite{liu2019roberta}. We describe both sets of features below:

\vspace{3mm} \noindent \textbf{Bag-of-Words.} We use TF-IDF-weighted unigrams and bigrams~\cite{schutze2008introduction} to build our content features for both customer and the brand's response posts. Specifically, we make use of content features from both the \textbf{Customer Post} and \textbf{Brand Agent Post}. We experiment with using either the initial customer or brand posts independently. Likewise, we evaluate the performance of using both posts. Note, when combing the posts, we treat the features from each group independently, e.g., there are two features for the word ``great,'' one for the customer post and one for the brand post. 

\vspace{3mm} \noindent \textbf{RoBERTA.} We also experiment with the RoBERTa model~\cite{liu2019roberta}. Ideally, we hope that our engineered features can match the performance of a complex neural network-based method, while also resulting in an interpretable model. Hence,  we compare with RoBERTa which is a strong baseline for many text classification tasks. Specifically, we experiment with using both the initial customer post $P = [w_1, \ldots, w_N]$ and the initial brand post $B = [w_1, \ldots, w_T]$ where $w_i$ represents word $i$. We evaluate the performance of each independently, along with combining them where we concatenate the brand post to the end of the customer post before processing with RoBERTa. The last layer's CLS token is passed to a final output layer for prediction.

\begin{table*}[t]
\centering
\resizebox{.8\textwidth}{!}{%
\renewcommand{\arraystretch}{1.1}
\begin{tabular}{lp{13cm}}
\toprule
 \textbf{Tags} & \textbf{Examples}                                                                                 \\ \midrule
Statement     & Updating to windows made it sooo   slow. Now the recent update killed it.                \\
Question      & Is this something you're seeing   now? Let us know in a DM and we'll take it from there. \\
Appreciation  & Thanks I've since had an email   from XXX and it's been sorted.                          \\
Response      & Not a problem XX, I hope your   future journeys are better. ZZZ.                         \\
Suggestion & That's good to hear! Would you   mind sharing your experience with us? Help us improve our support by   answering our survey here: Have a wonderful day! \\ \bottomrule
\end{tabular}%
}
\caption{Sample dialogue tags.}\vspace{0em}
\label{tab:sample_dialogue}
\end{table*}

\begin{table}[t]
\centering
\resizebox{0.7\linewidth}{!}{%
\begin{tabular}{lrrrr}
\toprule
         & \multicolumn{2}{c}{\textbf{BC3}}        & \multicolumn{2}{c}{\textbf{QC3}}        \\ \cmidrule(lr){2-3} \cmidrule(lr){4-5}
         
         & \textbf{BERT} & \textbf{JM} & \textbf{BERT} & \textbf{JM}  \\ 
         \toprule
\textbf{Accuracy} & .89 & .85                   & .87 & .78                   \\
\textbf{Macro F1} & .67 & .63                   & .70 & .65                   \\ \bottomrule
\end{tabular}}
\caption{Comparison of BERT to \citet{joty2018modeling} on the BC3 and QC3 dialogue acts datasets.}\vspace{0em}
\label{tab:DialogueActs}
\end{table}

\subsection{Stylistic Features}
As previously mentioned, in this study, we analyze both content and stylistic features. Content features measure what the text is about. Style features measure how it is written. This can include information about its author, its purpose, feelings it is meant to evoke, and more~\cite{argamon2007stylistic}. Hence, to construct the stylistic features of the discourse, we examine the expressed empathy of the response posts, the psycho-linguistic features of both customer and brand response posts, the dialogue tags, and the perplexity (uniqueness) of the brand's response posts. Each set of features is described below.

\vspace{3mm} \noindent \textbf{Empathy Identification Model}
For empathy identification, we use the framework introduced by \citet{sharma2020computational}. It consists of three communication mechanisms providing a comprehensive outlook of empathy---\textit{Emotional Reactions}, \textit{Interpretations}, and \textit{Explorations}. \textit{Emotional reactions} involve detecting texts related to emotions such as warmth, compassion, and concern, expressed by the brand agent after hearing about the customers' issue (e.g., \textit{I'm sorry you are having this problem}). \textit{Interpretations} involve the brand agent communicating a real understanding of the feelings and issues of the customer (e.g., \textit{I have also had t his issue before, I'm sorry, it really is annoying}). \textit{Explorations} involve the brand agent actively seeking/exploring the experience of the customer (e.g., \textit{what happened when you restarted your computer?}).
\begin{figure}[t]
    \centering
    \includegraphics[width=\linewidth]{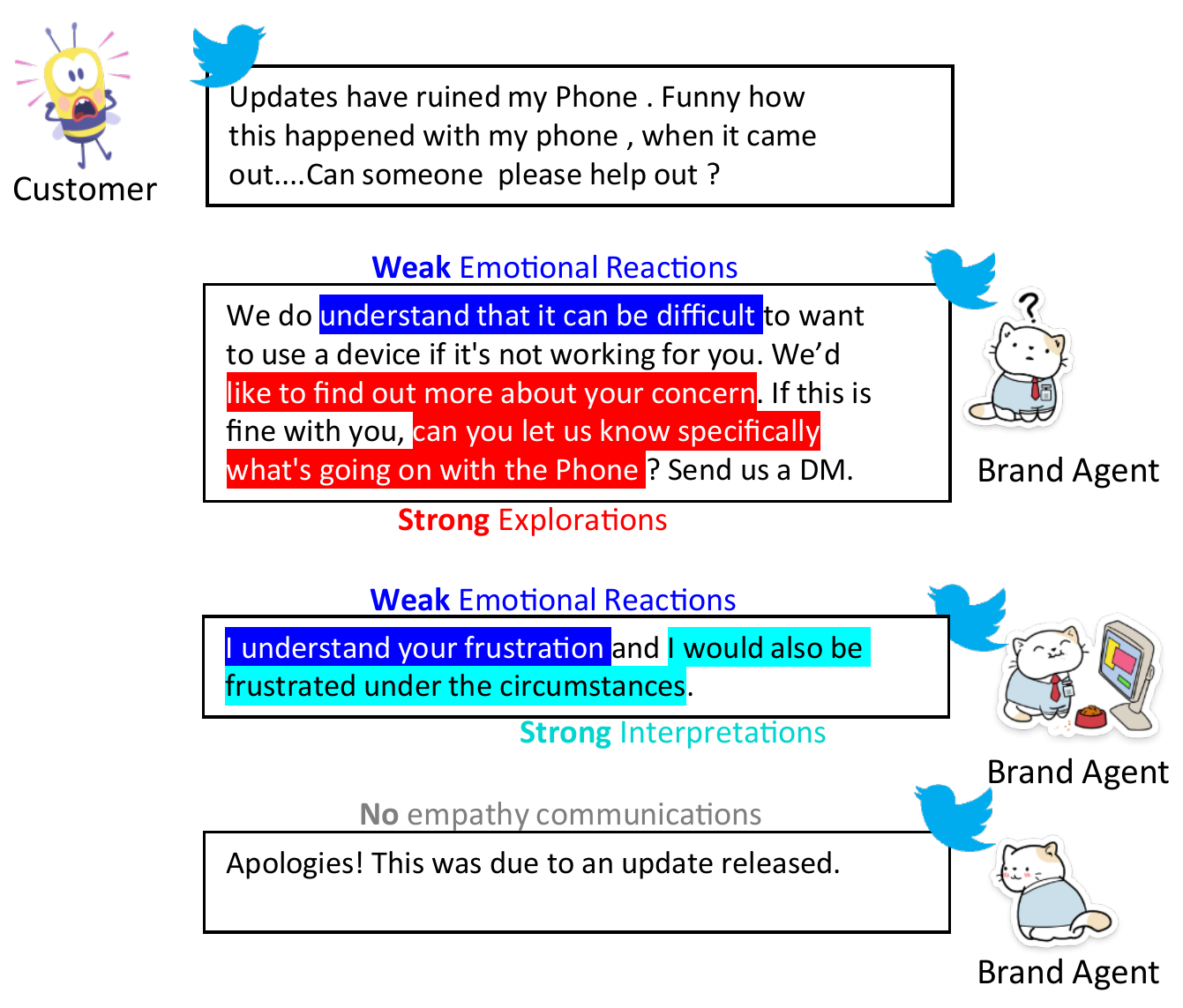}
    \caption{Examples of the three empathy communication mechanisms: Emotional Reactions, Interpretations, and Explorations based upon~\cite{sharma2020computational}. We differentiate between no communication, weak communication, and strong communication of these factors.}\vspace{0em}
    \label{fig:empathy}
\end{figure}
For each of these mechanisms, the study differentiates between, (0) no expression of empathy (no communication), (1) weak expression of empathy (weak communication), (2) strong expression of empathy (strong communication). To get these scores for each empathy mechanism, we use the pre-trained RoBERTa-based model by \citet{sharma2020computational}. The model leverages attention between neural representations of the seeker and response posts for generating a seeker-context aware representation of the response post, used to perform the two tasks of empathy identification and rationale extraction. We leverage this model and train it on the Reddit corpus of the \citet{sharma2020computational} dataset for 4 epochs using a learning rate of 2e-5, batch size of 32, $\lambda$EI $=$ 1, and $\lambda$RE $=$ .5. Figure~\ref{fig:empathy} provides sample responses to represent the three mechanisms of empathy. Someone can argue about the bias the training data set might introduce, given it is not related to customer support. However, we believe that the context is quite similar and there are several similarities that should minimize any bias. First, both data sets have a similar structure. A seeker who is seeking some answer or issue resolution that has been bothering him from an agent responsible for providing the response. Second, both data sets are online text-based communications, thus minimizing the bias again. We believe ``Empathy'' is a very universal thing and should not differ much with text-based communications. To evaluate this, we sampled a small set of tweets from our dataset and qualitatively found the model reliable. The examples provided in Figure~\ref{fig:empathy} are slight variants of tweets within our dataset that were correctly identified.

\vspace{3mm} \noindent \textbf{Perplexity.}
How can we measure whether a brand agent's response is generic or not? To do this, we use custom-built language models and measure their perplexity on each tweet. Specifically, we train a KenLM~\cite{heafield-2011-kenlm} n-gram-based language model on a held-out set of responses across all brands. We then use the language model to  calculate the probability of a response generated by the agent. We use the perplexity metric to score the response (probabilities), which is a commonly used metric for measuring the performance of a language model. Perplexity is defined as the inverse of the probability of
the test set normalized by the number of words
\begin{equation*}
 PPL(X) =  \sqrt{\prod_{i=1}^{N}P(w_i|w_{i-1})^{-\frac{1}{N}}}
\end{equation*}
where $P(w_i|w_{i-1})$ represents the probability of a word given the previous word and $N$ is the total number of words in a given agent's response. The equation above is an example using only ngrams. We train the KenLM model with a combination of 3-, 4-, and 5-grams.

Intuitively, another way of interpreting perplexity is the measure of the likelihood of a given test sentence in reference to the training corpus. Based on this intuition, we hypothesize the following: ``When a language model is primed with a collection of response tweets, the perplexity can serve as an indicator for personalization of a given brand's response.''
The rationale behind this is that the most common tweets would share more similarities (e.g., common terms and language patterns) with each other. This leads to common responses such ``\textit{How may I help you?}'' to have lower perplexity while unique responses such as ``\textit{Sorry to hear that! What is the exact version of the OS you're running and we'll figure out our next steps there. Thanks.}'' will have a higher perplexity score.  This hypothesis is supported by the use of perplexity to measure ``surprisal'' of misinformation and fake news when primed on factual knowledge~\cite{lee2021towards}.

\vspace{3mm} \noindent \textbf{Dialogue Tags.}
When people interact on social media, they interact with each other
at different times, performing certain communicative acts, called speech acts (e.g., question, request, statement). We hypothesize that the types of communicative acts made by the user and brand agent can have an impact on overall engagement. Table~\ref{tab:sample_dialogue} provides examples for dialogue tags.
To perform deep conversational analysis, we fine-tune a transformer model BERT~\cite{devlin2018bert} on the QC3~\cite{joty2018modeling} and BC3~\cite{ulrich2008publicly} data sets for speech act recognition and achieve superior performance than  original study~\cite{joty2018modeling}. We then use our model for qc3---based on the overall performance of their respective datasets and simple qualitative analysis of our data---to score dialogue tags for the initial posts for both customers and brand agents. Since, the speech acts identify the sentence structure as a question, suggestion, statement, appreciation, or response, which is more specific to linguistic structure than domain, we believe the bias introduced through the training data set would be minimal. Table~\ref{tab:DialogueActs} compares the results for our model. The predicted tags for each item is tweet is used as a feature in our final model. 


\vspace{3mm} \noindent \textbf{Psycho-Linguistic Features.}
To examine the language more deeply, we leverage the psycho-linguistic resources from the Linguistic Inquiry and Word Count (LIWC) tool~\cite{pennebaker2001linguistic,pennebaker2007linguistic} which has been psycho-metrically validated and performs well on social media data sets to extract lexico-syntactic features~\cite{de2013predicting}. LIWC provides a set of lexicons (word lists) for studying the various emotional, cognitive, and structural components present in individuals' written text. We extract several linguistic measures from LIWC, including the word count, psychological, cognitive, perceptual processes, and time orientation separately for customer and brand posts. We run each post independently through LIWC to generate independent scores.

\vspace{.25em}

\begin{table}[t]
\centering
\resizebox{\linewidth}{!}{%
\begin{tabular}{lrrr}
\toprule
\textbf{Model}                                                                & \textbf{Macro P.} & \textbf{Macro R.} & \textbf{Macro F1} \\ \midrule
Stratefied Baseline                                                  & .50             & .50          & .50      \\
Uniform Baseline                                                     & .50           & .50        & .41     \\
Minor Class baseline                         & .06          & .50          & .10    \\ \midrule

\multicolumn{4}{c}{RoBERTa Models} \\ \midrule
RoBERTa + Customer Post                                              & .59            & .57         & .58    \\
RoBERTa + BAP                                        & .73            & .72         & .72      \\
RoBERTa + CP +   BAP                        & \textbf{.73}            & {.73}         &\textbf{.73} \\ \midrule
\multicolumn{4}{c}{Linear BoW Models} \\ \midrule
CP                                                      & .58            & .61         & .58      \\
BAP & .65            & .77         & .67     \\
CP + BAP & .65            & .75         & .68    \\
LIWC + E + P + DT & .57	& .68        & .54      \\
CP+BAP+LIWC+E+P+DT & .69 &	\textbf{.76}       & \textbf{.72}    \\ \bottomrule
\end{tabular}%
}
\caption{Main Results for different feature sets: Customer Post (CP), Brand Agent Post (BAP), Perplexity (P), Dialogue Tags (DT), and Empathy (E).}
\vspace{0em}
\label{tab:results}
\end{table}

\begin{table}[t]
\centering
\resizebox{\linewidth}{!}{%
\begin{tabular}{lrrrrr}
\toprule
 & \textbf{Macro P.} & \textbf{Macro R.} & \textbf{Macro F1}  \\ \midrule
CP +  BAP + LIWC + E + P+ DA & .69             & .77          & .72       \\
-- perplexity   (P)                                                       & .60             & .73          & .57       \\
-- empathy (E)                                                         & .60             & .73          & .58      \\
-- LIWC                                                                & .65             & .77          & .66      \\
-- Dialogue Acts (DA)                                                                & .66             & .78          & .69     \\
-- Brand Agent Post  (BAP)                                                  & .62             & .65          & .63    \\
-- Customer Post  (CP)                                                     & .67             & .80          & .69     \\ \bottomrule
\end{tabular}%
}
\caption{Ablation Results for different feature sets using the linear model: Customer Post (CP), Brand Agent Post (BAP), Perplexity (P), Dialogue Tags (DT), and Empathy (E).}\vspace{0em}
\label{tab:ablation}
\end{table}

\section{Experiments}
This section evaluates how style and content features impact the general predictive performance of the machine learning models.

\vspace{3mm}
\noindent \textbf{Model Training Details.}
For classification, we used the dichotomous dependent variable: ``Engagement Indicator''. Specifically, using the features described in the previous section, we train a classification model. The classification model is trained to predict the ``engagement'' class, i.e., whether the length of the customer$\rightarrow$brand responses is greater than one.  We train the Logistic regression model using the scikit-learn package~\cite{pedregosa2011scikit}. Additionally, we also trained a transformer-based model-RoBERTa with binary cross-entropy classification loss. We trained this model using a learning rate of 2e-5 and a batch size of 32 on 4 epochs. Hyperparameters for all models were chosen using a randomly sampled validation partition of 10\% of the training data.

\vspace{3mm}
\noindent \textbf{Baselines.}
We report the results of various baselines. Specifically, we compare various naive baselines, including three random classification baselines: Stratified, Minor Class, and Uniform for classification.
Stratified randomly generates predictions based on the class (Engaging and Not-Engaging) proportions. Uniform randomly generates predictions equally for both classes, independent of the class frequency. The Minor Class baseline always predicts the least frequent class. Beyond the naive baselines, we compare three models using content features: Logistic Regression models trained on the Customer Post, Brand Agent Post, and Customer + Brand Agent Posts. All models using content features make use of TF-IDF-weighted unigram and bigram features. Furthermore, we compare to a model that uses the stylistic features for customers and brand agents (LIWC + Empathy + Perplexity + Dialogue Tags), both independently and combined. Finally, ``our'' method uses all of the content and style features across brands and customers (Customer Post +  Brand Agent Post + LIWC + Empathy +   Perplexity + Dialogue Tags).

\vspace{3mm}
\noindent \textbf{Results.}
Table~\ref{tab:results} reports the Macro Precision, Macro Recall, and Macro F1. We make two major observations. First, we find that our method outperforms the naive (random) baselines. The Macro-F1 score for the naive baselines is highest for the Stratified Baseline with an F1 of .50. On the contrary, the Minor class Baseline, performs poorly for the Macro F1 score (.10), i.e., because it always predicts the ``non-engagement'' class. Next, we find that Brand Post content features are more predictive than the customer's original post. Hence, the Brand's response is vital for promoting engagement, more so than the original customer's tweet.
Finally, the combination of content features and stylistic features performs best with a Macro  of .72 almost matching the best Macro F1 (0.73) for the more complex (uninterpretable) RoBERTa model. Moreover, the combination method outperforms all other methods with regard to Macro Recall, suggesting that the linear models are more robust using all of the engineered features. See the Appendix for implications and further discussion about the results.


\vspace{3mm}
\noindent \textbf{Ablation Study.}
Next, we analyze the components in our classification models through an ablation study for the model using all the features. Intuitively, we wish to test which set of features has the largest impact on model performance. Table~\ref{tab:ablation} summarizes our findings. Interestingly, we see the most significant drops in performance from removing Perplexity and Empathy information (e.g., removing perplexity features drops the Macro F1 from .72 to .57, and removing empathy drops it to .58), indicating complex relationships with the other features in the classification model. 

\begin{table}[t]
\centering
\resizebox{\linewidth}{!}{%
\begin{tabular}{llr}
\toprule
\textbf{Feature Group}              & \textbf{Feature}                 & \textbf{Importance} \\ \midrule
\multirow{3}{*}{\textbf{Empathy}}        & Exploratorations        & .126      \\
                                & Interpretations         & -.004     \\
                                & Emotional Reactions     & -.034     \\ \midrule
\multirow{10}{*}{\textbf{Dialogue Tags}} & BRAND: questions        & .072      \\
                                & CUSTOMER: statements    & .019      \\
                                & CUSTOMER: response      & .012      \\
                                & BRAND: suggestions      & .007      \\ \cmidrule(lr){2-3}
                                & CUSTOMER: appreciations & -.003     \\
                                & CUSTOMER: suggestions   & -.006     \\
                                & CUSTOMER: questions     & -.019     \\
                                & BRAND: statements       & -.042     \\
                                & BRAND: appreciations    & -.044     \\
                                & BRAND: response         & -.049     \\ \midrule
\multirow{16}{*}{\textbf{LIWC}}          & CUSTOMER: word\_count   & .136      \\
                                & BRAND: word\_count      & .116      \\
                                & BRAND: interrogation    & .092      \\
                                & CUSTOMER: time          & .089      \\
                                & BRAND: time             & .088      \\
                                & BRAND: focuspast        & .061      \\
                                & BRAND: focuspresent     & .041      \\
                                & BRAND: certain          &
                                .040      \\ \cmidrule(lr){2-3}
                                & CUSTOMER: tentative     & -.014     \\
                                & BRAND: informal         & -.015     \\
                                & CUSTOMER: informal      & -.016     \\
                                & CUSTOMER: focusfuture   & -.017     \\
                                & CUSTOMER: focuspast     & -.035     \\
                                & BRAND: focusfuture      & -.055     \\
                                & BRAND: insight          & -.114     \\
                                & BRAND: Tone             & -.139     \\
                                & BRAND: Clout            & -.319     \\ \midrule
\textbf{Personalization}                 & BRAND: Novelty          & -.026     \\ \bottomrule
\end{tabular}%
}
\caption{Feature Importance for Stylistic Features used to predict engagement}\vspace{0em}
\label{tab:feature_imp}
\end{table}

\vspace{3mm}
\noindent \textbf{Feature Importance.}
Next, we perform a comprehensive analysis of our model focusing on the coefficient scores of the logistic regression model to analyze individual feature impact on model performance. Our paper reveals several insights for brand agents as well as customers. Table~\ref{tab:feature_imp} summarizes our feature importance results for features with the largest magnitudes (positive and negative). At a high level, we find that Empathy Explorations are of substantial importance for positive engagement. Likewise, the LIWC category Clout indicates negative relationships for engagement. This is interesting because a Higher Clout score is marked by using more we-words and social words and fewer I-words and
negations (e.g., no, not). This indicates that users engage more when brands take responsibility for issues (e.g., ``I will find you a solution'' vs. ``we can work together to fix it''). This is further supported by the positive relationship for words with high certainty made used by the Brand (i.e., BRAND: certain). Lastly, we find that Novelty has a small negative coefficient score. After further analysis, we find that when there is a slight chance that a highly novel initial response by the brand can quickly solve the problem right away, limiting the need for further discussion---which is a good thing. However, this is somewhat rare in our analysis. The overall recommendations based on our findings are summarized in Figure~\ref{fig:my_label}. We summarize the key implications/recommendations in the following subsections.

\section{Conclusion}
Our study demonstrates that even though some customer support requests on social media might reflect anger, there are some key indicators that can engage customers in a positive direction. Most extant research has not paid any significant attention to this aspect of the discourse, mostly simply focusing on sentiment or timely response. Hence, we examined text based, asynchronous social media discourses for both \textit{what is written}  and \textit{how it is written}, to examine how these features influence customer engagement. Our study effectively identifies multiple such stylistic features that can influence the engagement of these social discourses between customers and brands. 
\section{Limitations and Future Research}
A limitation of our study is that we proxy engagement with the length or the number of rounds customers and brands respond to each other on the social platform and assume that customers appreciate the continuous effort from the brand to resolve their issues, even when they are not resolved. Our initial analysis supports this, however, future research can validate the perceived effort through a natural field experiment or a lab setting. Moreover, we do not account for engagement outside the social platform. Another limitation of our paper is that it is exploratory and based on observational data. A controlled lab experiment studying the sentiment of the conversations along with the stylistic/linguistic features to contrast the two aspects and support the claim empirically can establish the claims further. Furthermore, even when this is not the complete case, understanding what gets responses is important, even if it is to point agents towards what to avoid. Also, while we believe engagement is a proxy for quality interactions from the customers' perspective (in general), engagement is not a good thing in all cases from a customer service perspective because it increases the time agents are working with each customer on average. However, these results are also useful for future research in customer service chatbot development, which can be useful to develop bots that show direct care for customers. Moreover, while it can increase the cost of customer service, social media is also acting as a strong marketing tool, which can increase revenue for the company, hence, the limitation may not be as strong as potentially expected; however, this would need to be tested.

\section*{Acknowledgements}

This material is based upon work supported by the National Science Foundation (NSF) under Grant~No. 2145357.

\bibliography{references}

\begin{thebibliography}{41}
\expandafter\ifx\csname natexlab\endcsname\relax\def\natexlab#1{#1}\fi

\bibitem[{Adam et~al.(2020)Adam, Wessel, and Benlian}]{adam2020ai}
Martin Adam, Michael Wessel, and Alexander Benlian. 2020.
\newblock Ai-based chatbots in customer service and their effects on user
  compliance.
\newblock \emph{Electronic Markets}, pages 1--19.

\bibitem[{Argamon et~al.(2007)Argamon, Whitelaw, Chase, Hota, Garg, and
  Levitan}]{argamon2007stylistic}
Shlomo Argamon, Casey Whitelaw, Paul Chase, Sobhan~Raj Hota, Navendu Garg, and
  Shlomo Levitan. 2007.
\newblock Stylistic text classification using functional lexical features.
\newblock \emph{Journal of the American Society for Information Science and
  Technology}, 58(6):802--822.

\bibitem[{Bargh and McKenna(2004)}]{bargh2004internet}
John~A Bargh and Katelyn~YA McKenna. 2004.
\newblock The internet and social life.
\newblock \emph{Annu. Rev. Psychol.}, 55:573--590.

\bibitem[{Berger et~al.(2020)Berger, Humphreys, Ludwig, Moe, Netzer, and
  Schweidel}]{berger2020uniting}
Jonah Berger, Ashlee Humphreys, Stephan Ludwig, Wendy~W Moe, Oded Netzer, and
  David~A Schweidel. 2020.
\newblock Uniting the tribes: Using text for marketing insight.
\newblock \emph{Journal of Marketing}, 84(1):1--25.

\bibitem[{Bhuiyan et~al.(2018)Bhuiyan, Misra, Tripathy, Mahmud, and
  Akkiraju}]{bhuiyan2018don}
Mansurul Bhuiyan, Amita Misra, Saurabh Tripathy, Jalal Mahmud, and Rama
  Akkiraju. 2018.
\newblock Don't get lost in negation: An effective negation handled dialogue
  acts prediction algorithm for twitter customer service conversations.
\newblock \emph{arXiv preprint arXiv:1807.06107}.

\bibitem[{Brown et~al.(1992)Brown, Della~Pietra, Della~Pietra, Lai, and
  Mercer}]{brown1992estimate}
Peter~F Brown, Stephen~A Della~Pietra, Vincent~J Della~Pietra, Jennifer~C Lai,
  and Robert~L Mercer. 1992.
\newblock An estimate of an upper bound for the entropy of english.
\newblock \emph{Computational Linguistics}, 18(1):31--40.

\bibitem[{Clark et~al.(2013)Clark, Murfett, Rogers, and Ang}]{clark2013empathy}
Colin~Mackinnon Clark, Ulrike~Marianne Murfett, Priscilla~S Rogers, and Soon
  Ang. 2013.
\newblock Is empathy effective for customer service? evidence from call center
  interactions.
\newblock \emph{Journal of Business and Technical Communication},
  27(2):123--153.

\bibitem[{De~Choudhury et~al.(2013)De~Choudhury, Gamon, Counts, and
  Horvitz}]{de2013predicting}
Munmun De~Choudhury, Michael Gamon, Scott Counts, and Eric Horvitz. 2013.
\newblock Predicting depression via social media.
\newblock In \emph{Proceedings of ICWSM}, volume~7.

\bibitem[{Devlin et~al.(2018)Devlin, Chang, Lee, and
  Toutanova}]{devlin2018bert}
Jacob Devlin, Ming-Wei Chang, Kenton Lee, and Kristina Toutanova. 2018.
\newblock Bert: Pre-training of deep bidirectional transformers for language
  understanding.
\newblock \emph{arXiv preprint arXiv:1810.04805}.

\bibitem[{F{\o}lstad and Taylor(2019)}]{folstad2019conversational}
Asbj{\o}rn F{\o}lstad and Cameron Taylor. 2019.
\newblock Conversational repair in chatbots for customer service: The effect of
  expressing uncertainty and suggesting alternatives.
\newblock In \emph{International Workshop on Chatbot Research and Design},
  pages 201--214. Springer.

\bibitem[{Forbes(2018)}]{Forbes}
Forbes. 2018.
\newblock Forbes.
\newblock
  \url{https://www.forbes.com/sites/shephyken/2018/08/05/what-customers-want-and-expect/#5a4c6f7a7701}.
\newblock Accessed: 2021-01-01.

\bibitem[{Greenberg(2010)}]{greenberg2010crm}
Paul Greenberg. 2010.
\newblock \emph{CRM at the speed of light: Social CRM strategies, tools, and
  techniques}.
\newblock McGraw-Hill New York.

\bibitem[{Guercini et~al.(2014)Guercini, Misopoulos, Mitic, Kapoulas, and
  Karapiperis}]{guercini2014uncovering}
Simone Guercini, Fotis Misopoulos, Miljana Mitic, Alexandros Kapoulas, and
  Christos Karapiperis. 2014.
\newblock Uncovering customer service experiences with twitter: the case of
  airline industry.
\newblock \emph{Management Decision}.

\bibitem[{Hancock and Dunham(2001)}]{hancock2001impression}
Jeffrey~T Hancock and Philip~J Dunham. 2001.
\newblock Impression formation in computer-mediated communication revisited: An
  analysis of the breadth and intensity of impressions.
\newblock \emph{Communication research}, 28(3):325--347.

\bibitem[{Heafield(2011)}]{heafield-2011-kenlm}
Kenneth Heafield. 2011.
\newblock {K}en{LM}: Faster and smaller language model queries.
\newblock In \emph{Proceedings of the Sixth Workshop on Statistical Machine
  Translation}, pages 187--197.

\bibitem[{Herzig et~al.(2016)Herzig, Feigenblat, Shmueli-Scheuer, Konopnicki,
  Rafaeli, Altman, and Spivak}]{herzig-etal-2016-classifying}
Jonathan Herzig, Guy Feigenblat, Michal Shmueli-Scheuer, David Konopnicki, Anat
  Rafaeli, Daniel Altman, and David Spivak. 2016.
\newblock \href {https://doi.org/10.18653/v1/W16-3609} {Classifying emotions in
  customer support dialogues in social media}.
\newblock In \emph{Proceedings of SIGdial}, pages 64--73, Los Angeles.
  Association for Computational Linguistics.

\bibitem[{Hu et~al.(2018)Hu, Xu, Liu, You, Guo, Sinha, Luo, and
  Akkiraju}]{hu2018touch}
Tianran Hu, Anbang Xu, Zhe Liu, Quanzeng You, Yufan Guo, Vibha Sinha, Jiebo
  Luo, and Rama Akkiraju. 2018.
\newblock Touch your heart: A tone-aware chatbot for customer care on social
  media.
\newblock In \emph{Proceedings of CHI}, pages 1--12.

\bibitem[{Jacobson(1999)}]{jacobson1999impression}
David Jacobson. 1999.
\newblock Impression formation in cyberspace: Online expectations and offline
  experiences in text-based virtual communities.
\newblock \emph{Journal of Computer-Mediated Communication}, 5(1):JCMC511.

\bibitem[{Joty and Mohiuddin(2018)}]{joty2018modeling}
Shafiq Joty and Tasnim Mohiuddin. 2018.
\newblock Modeling speech acts in asynchronous conversations: A neural-crf
  approach.
\newblock \emph{Computational Linguistics}, 44(4):859--894.

\bibitem[{Lee et~al.(2021)Lee, Bang, Madotto, and Fung}]{lee2021towards}
Nayeon Lee, Yejin Bang, Andrea Madotto, and Pascale Fung. 2021.
\newblock Towards few-shot fact-checking via perplexity.
\newblock In \emph{Proceedings of the 2021 Conference of the North American
  Chapter of the Association for Computational Linguistics: Human Language
  Technologies}, pages 1971--1981.

\bibitem[{Liu et~al.(2020)Liu, Gao, Kang, Jiang, He, Sun, Liu, and
  Lu}]{liu2020time}
Jiawei Liu, Zhe Gao, Yangyang Kang, Zhuoren Jiang, Guoxiu He, Changlong Sun,
  Xiaozhong Liu, and Wei Lu. 2020.
\newblock Time to transfer: Predicting and evaluating machine-human chatting
  handoff.
\newblock \emph{arXiv preprint arXiv:2012.07610}.

\bibitem[{Liu et~al.(2019)Liu, Ott, Goyal, Du, Joshi, Chen, Levy, Lewis,
  Zettlemoyer, and Stoyanov}]{liu2019roberta}
Yinhan Liu, Myle Ott, Naman Goyal, Jingfei Du, Mandar Joshi, Danqi Chen, Omer
  Levy, Mike Lewis, Luke Zettlemoyer, and Veselin Stoyanov. 2019.
\newblock Roberta: A robustly optimized bert pretraining approach.
\newblock \emph{arXiv preprint arXiv:1907.11692}.

\bibitem[{Mousavi et~al.(2020)Mousavi, Johar, and Mookerjee}]{mousavi2020voice}
Reza Mousavi, Monica Johar, and Vijay~S Mookerjee. 2020.
\newblock The voice of the customer: Managing customer care in twitter.
\newblock \emph{Information Systems Research}, 31(2):340--360.

\bibitem[{Oraby et~al.(2017)Oraby, Gundecha, Mahmud, Bhuiyan, and
  Akkiraju}]{oraby2017may}
Shereen Oraby, Pritam Gundecha, Jalal Mahmud, Mansurul Bhuiyan, and Rama
  Akkiraju. 2017.
\newblock " how may i help you?" modeling twitter customer service
  conversations using fine-grained dialogue acts.
\newblock In \emph{Proceedings of the IUI}, pages 343--355.

\bibitem[{Packard and Berger(2021)}]{packard2021concrete}
Grant Packard and Jonah Berger. 2021.
\newblock How concrete language shapes customer satisfaction.
\newblock \emph{Journal of Consumer Research}, 47(5):787--806.

\bibitem[{Packard et~al.(2018)Packard, Moore, and McFerran}]{packard2018m}
Grant Packard, Sarah~G Moore, and Brent McFerran. 2018.
\newblock (i'm) happy to help (you): The impact of personal pronoun use in
  customer--firm interactions.
\newblock \emph{Journal of Marketing Research}, 55(4):541--555.

\bibitem[{Pedregosa et~al.(2011)Pedregosa, Varoquaux, Gramfort, Michel,
  Thirion, Grisel, Blondel, Prettenhofer, Weiss, Dubourg
  et~al.}]{pedregosa2011scikit}
Fabian Pedregosa, Ga{\"e}l Varoquaux, Alexandre Gramfort, Vincent Michel,
  Bertrand Thirion, Olivier Grisel, Mathieu Blondel, Peter Prettenhofer, Ron
  Weiss, Vincent Dubourg, et~al. 2011.
\newblock Scikit-learn: Machine learning in python.
\newblock \emph{the Journal of machine Learning research}, 12:2825--2830.

\bibitem[{Pennebaker(2011)}]{pennebaker2011using}
James~W Pennebaker. 2011.
\newblock Using computer analyses to identify language style and aggressive
  intent: The secret life of function words.
\newblock \emph{Dynamics of Asymmetric Conflict}, 4(2):92--102.

\bibitem[{Pennebaker et~al.(2007)Pennebaker, Booth, and
  Francis}]{pennebaker2007linguistic}
James~W Pennebaker, Roger~J Booth, and Martha~E Francis. 2007.
\newblock Linguistic inquiry and word count: Liwc [computer software].
\newblock \emph{Austin, TX: liwc. net}, 135.

\bibitem[{Pennebaker et~al.(2015)Pennebaker, Boyd, Jordan, and
  Blackburn}]{pennebaker2015development}
James~W Pennebaker, Ryan~L Boyd, Kayla Jordan, and Kate Blackburn. 2015.
\newblock The development and psychometric properties of liwc2015.
\newblock Technical report.

\bibitem[{Pennebaker et~al.(2001)Pennebaker, Francis, and
  Booth}]{pennebaker2001linguistic}
James~W Pennebaker, Martha~E Francis, and Roger~J Booth. 2001.
\newblock Linguistic inquiry and word count: Liwc 2001.
\newblock \emph{Mahway: Lawrence Erlbaum Associates}, 71(2001):2001.

\bibitem[{Power(2013)}]{Power2013}
Power. 2013.
\newblock Power 2013.
\newblock
  \url{https://www.jdpower.com/business/press-releases/2013-social-media-benchmark-study}.
\newblock Accessed: 2021-01-01.

\bibitem[{Rouse and Haas(2003)}]{rouse2003exploring}
Steven~V Rouse and Heather~A Haas. 2003.
\newblock Exploring the accuracies and inaccuracies of personality perception
  following internet-mediated communication.
\newblock \emph{Journal of research in personality}, 37(5):446--467.

\bibitem[{Sch{\"u}tze et~al.(2008)Sch{\"u}tze, Manning, and
  Raghavan}]{schutze2008introduction}
Hinrich Sch{\"u}tze, Christopher~D Manning, and Prabhakar Raghavan. 2008.
\newblock \emph{Introduction to information retrieval}, volume~39.
\newblock Cambridge University Press Cambridge.

\bibitem[{Sharma et~al.(2020)Sharma, Miner, Atkins, and
  Althoff}]{sharma2020computational}
Ashish Sharma, Adam~S Miner, David~C Atkins, and Tim Althoff. 2020.
\newblock A computational approach to understanding empathy expressed in
  text-based mental health support.
\newblock \emph{arXiv preprint arXiv:2009.08441}.

\bibitem[{Ulrich et~al.(2008)Ulrich, Murray, and Carenini}]{ulrich2008publicly}
Jan Ulrich, Gabriel Murray, and Giuseppe Carenini. 2008.
\newblock A publicly available annotated corpus for supervised email
  summarization.

\bibitem[{Wieseke et~al.(2012)Wieseke, Geigenm{\"u}ller, and
  Kraus}]{wieseke2012role}
Jan Wieseke, Anja Geigenm{\"u}ller, and Florian Kraus. 2012.
\newblock On the role of empathy in customer-employee interactions.
\newblock \emph{Journal of service research}, 15(3):316--331.

\bibitem[{Woodcock et~al.(2011)Woodcock, Green, and
  Starkey}]{woodcock2011social}
Neil Woodcock, Andrew Green, and Michael Starkey. 2011.
\newblock Social crm as a business strategy.
\newblock \emph{Journal of Database Marketing \& Customer Strategy Management},
  18(1):50--64.

\bibitem[{Xu et~al.(2017)Xu, Liu, Guo, Sinha, and Akkiraju}]{xu2017new}
Anbang Xu, Zhe Liu, Yufan Guo, Vibha Sinha, and Rama Akkiraju. 2017.
\newblock A new chatbot for customer service on social media.
\newblock In \emph{Proceedings of CHI}, pages 3506--3510.

\bibitem[{Yahav et~al.(2020)Yahav, Schwartz, and Welcman}]{yahav2020journey}
Inbal Yahav, David~G Schwartz, and Yaara Welcman. 2020.
\newblock The journey to engaged customer community: Evidential social crm
  maturity model in twitter.
\newblock \emph{Applied Stochastic Models in Business and Industry},
  36(3):397--416.

\bibitem[{Zhang et~al.(2011)Zhang, Erickson, and Webb}]{zhang2011effects}
Lu~Zhang, Lee~B Erickson, and Heidi~C Webb. 2011.
\newblock Effects of “emotional text” on online customer service chat.
\newblock \emph{Graduate Student Research Conference in Hospitality and
  Tourism}.

\end{thebibliography}
\bibliographystyle{acl_natbib}

\appendix

\section{Appendix}
\label{sec:appendix}

\subsection{Brand Agent Implications}
We examine both the emotional and cognitive aspects of expressed empathy and find that it can positively affect the likelihood of a successful interaction. Similarly, the level of personalization or novelty (measured using perplexity) of the brand agent's initial response and their time orientation also reveal some interesting insights for brand managers. The findings of our study provide new insights to brands for orchestrating an effective and engaging customer service discourse on social platforms, thus building great customer relationships. Additionally, our findings  can  also  provide  insights  into  improving  customer service chatbot responses by incorporating the stylistic features that we discuss in the study. We summarize some of the main findings that are relevant for brands below:


\vspace{3mm} \noindent \textbf{Expressing more  \textit{exploration} empathy about customer's issue in the initial response increases the likelihood of an engaging interaction.}
We find that the fact whether or not an agent's response communicates emotional (e.g., \textit{I'm sorry}) reactions might not be as important as  explorations (i.e., \textit{can you share the error message on your iPhone?}) for an engaging conversation indicating that exploring issues that a customer is facing influences the engagement of an interaction. Moreover, indicating generic understanding in form of interpretation empathy might help to resolve the issue quickly as it affects the length of the conversation negatively. This is also validated by other sets of stylistic features - Dialogue Tags (BRAND: questions and BRAND: suggestions) and LIWC (BRAND: interrogation) when brand agents ask more questions and provide more suggestions that lead to more engaging conversations by eliciting responses from customers.

\begin{figure*}[t]
    \centering
    \includegraphics[width=0.7\textwidth]{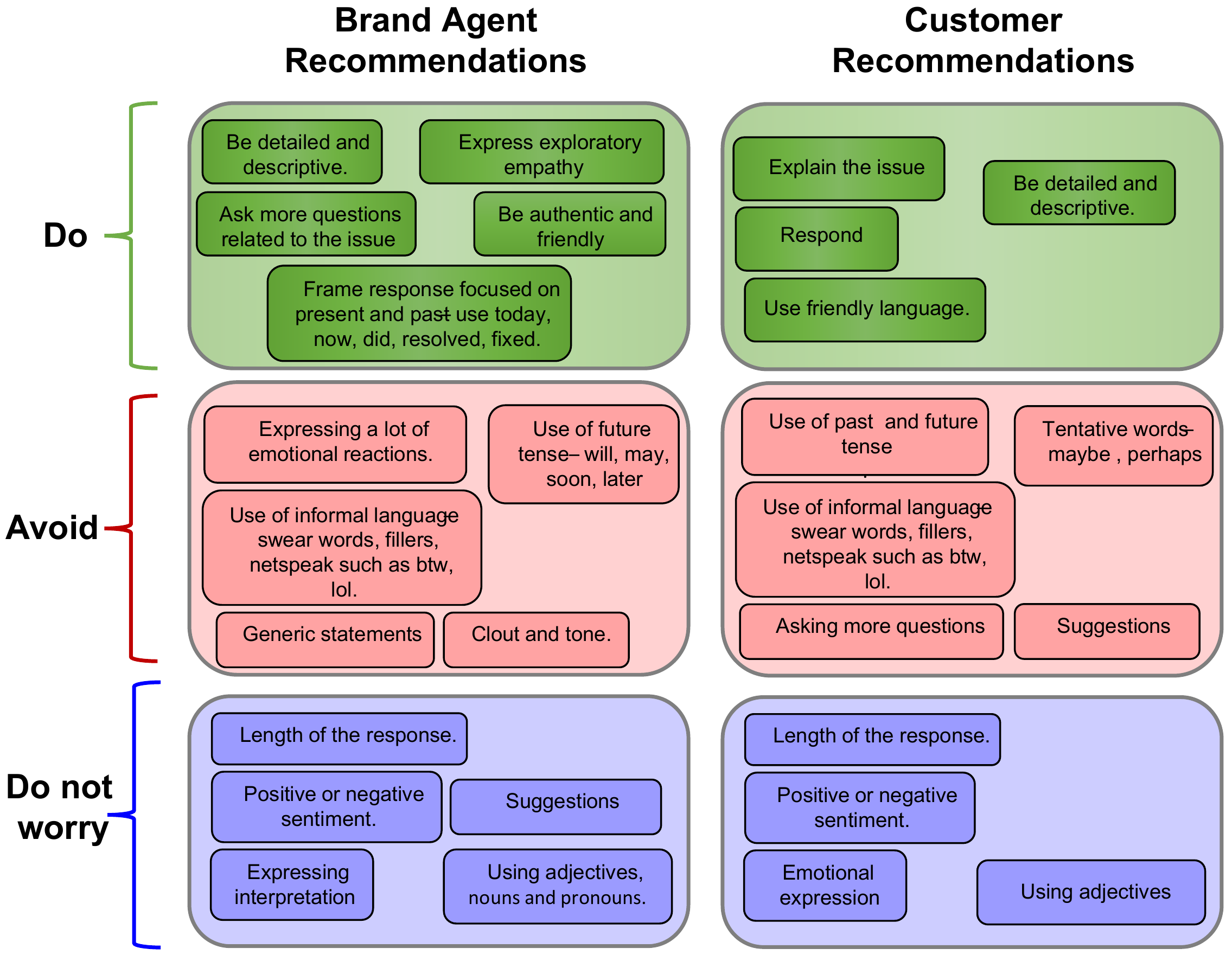}
    \caption{Overall recommendations for brand agents and customers based on this paper's findings.}
    \vspace{-1em}
    \label{fig:my_label}
\end{figure*}

\vspace{3mm} \noindent \textbf{Initial responses focused on the future from brand agent decreases the engagement of an interaction}
Brand agents often use phrases such as ``\textit{We will look into this}'' or ``\textit{We will get back to you}''. Our results show that the higher future orientation (\textit{LIWC-BRAND: focusfuture}) in the initial response post can lower the engagement of an interaction. An alternative solution based on our results could be to use exploration empathy to understand more about the customer's issue. On the other hand, the use of past-tense verbs (``\textit{I fixed it for you}'') and present focused (``\textit{We are looking into this now''} increase the engagement of an interaction. Thus, a general recommendation is to avoid making future promises, instead focus on responding when the incident is resolved, or state that you are actively working on it.

\subsection{Customers/Users Implications}
Our study also identifies some key implications for customers. The main intention of any customer reaching out to a brand on social media most typically is to get some issue resolved quickly. No matter how motivated or well-intentioned brand agents might be to resolve these issues, given the frequency and load of such issues they might leave many of such posts unattended or not provide satisfactory resolutions. We therefore also provide some guidelines to customers on how to increase the effectiveness of such interactions on social platforms.

\vspace{3mm} \noindent \textbf{Interrogative customer posts with informal language lower the engagement of an interaction } Our results show that the more interrogative and informal the initial customer post is (\textit{CUSTOMER: questions, CUSTOMER: informal}), the less likely it is going to be engaging. For instance, customer posts asking questions and using swear words or informal language are less likely to be engaging on the platform. This finding is substantial as this can help customers to frame their issues in a more explanatory manner rather than being informal and interrogative of the brands.

\vspace{3mm} \noindent \textbf{Customer posts focused too much in the past or future or tentative are less engaging}. We find that when the initial customer posts contain past-focused or future-focused words such as 'had it enough' 'will see you, or “may” or “perhaps it is likely to lead to an engaging interaction. This is an interesting finding because the usage of such words might signal that customer is not expecting their issue to be resolved and thus, brands might choose to attend to other posts, given the rate of customer service requests pouring on social media. 
For instance, given two customer posts ``\textit{May be someone could help with this issue?}'' and ``\textit{Please help to resolve this issue}'', the latter post signals the brand agent to take an action (and increases the likelihood of a success), while the former that signals tentative action.

\end{document}